\documentclass[letterpaper, 10 pt, conference]{ieeeconf}  

\IEEEoverridecommandlockouts                              

\usepackage{cite}
\usepackage{amsmath,amssymb,amsfonts}
\usepackage{amssymb}
\usepackage{pifont}
\usepackage{gensymb}
\usepackage[ruled]{algorithm2e}
\usepackage{algpseudocode}
\usepackage{graphicx}
\usepackage{textcomp}
\usepackage{siunitx}
\usepackage{multirow}
\usepackage{bbding}
\usepackage{booktabs}
\usepackage{lscape}
\usepackage{tablefootnote}
\usepackage{refcount}
\usepackage{threeparttablex}
\usepackage{kotex}
\usepackage{color, soul}
\usepackage[table,xcdraw]{xcolor}
\usepackage{hyperref}

\SetCommentSty{mycommfont}

\newcolumntype{?}[1]{!{\vrule width #1}}

\newcommand{\etal}{\textit{et al}. }
\newcommand*{\rom}[1]{\uppercase\expandafter{\romannumeral #1\relax}}
\newcolumntype{M}[1]{>{\centering\arraybackslash}m{#1}}

\title{\LARGE \bf mmWave Radar-Based Non-Line-of-Sight Pedestrian Localization at T-Junctions Utilizing Road Layout Extraction via Camera}

\author{Byeonggyu Park$^{1}$, Hee-Yeun Kim$^{1}$, Byonghyok Choi$^{2}$, Hansang Cho$^{2}$, Byungkwan Kim$^{3}$,\\Soomok Lee$^{4}$, Mingu Jeon$^{1}$, and Seong-Woo Kim$^{1}$
	\thanks{This work was supported by Samsung Electro-Mechanics Co., Ltd., the National Research Foundation of Korea (NRF) through the Ministry of Science and ICT under Grant 2021R1A2C1093957, Korea Institute for Advancement of Technology (KIAT) grant funded by the Korea Government (MOTIE) (P0020536, HRD Program for Industrial Innovation), the Korean Ministry of Land, Infrastructure and Transport (MOLIT) as the Innovative Talent Education Program for Smart City, and by the Institute of Engineering Research at Seoul National University, which provided the research facilities for this work.}
	\thanks{$^{1}$Seoul National University, $^{2}$Samsung Electro-Mechanics Co., Ltd., $^{3}$Chungnam National University, $^{4}$Ajou University. Correspondence to: Mingu Jeon and Seong-Woo Kim {\tt\small \{mingujeon; snwoo\}@snu.ac.kr}.}%
}

\begin{document}

\maketitle
\thispagestyle{empty}
\pagestyle{empty}


\begin{abstract}
Pedestrians Localization in Non-Line-of-Sight (NLoS) regions within urban environments poses a significant challenge for autonomous driving systems. While mmWave radar has demonstrated potential for detecting objects in such scenarios, the 2D radar point cloud (PCD) data is susceptible to distortions caused by multipath reflections, making accurate spatial inference difficult. Additionally, although camera images provide high-resolution visual information, they lack depth perception and cannot directly observe objects in NLoS regions. In this paper, we propose a novel framework that interprets radar PCD through road layout inferred from camera for localization of NLoS pedestrians. The proposed method leverages visual information from the camera to interpret 2D radar PCD, enabling spatial scene reconstruction. The effectiveness of the proposed approach is validated through experiments conducted using a radar-camera system mounted on a real vehicle. The localization performance is evaluated using a dataset collected in outdoor NLoS driving environments, demonstrating the practical applicability of the method.
\end{abstract}



\section{Introduction}\label{sec:Introduction}

Recent advancements in autonomous driving technology for robots and vehicles have been rapidly progressing and are being widely applied across various industrial sectors \cite{kim2017autonomous}. Currently, autonomous driving systems perceive the surrounding environment in real-time through various sensors, such as LiDAR, camera, and radar. In particular, LiDAR and camera are widely used for object localization and detection, operating based on Line-of-Sight (LoS) principles to recognize the visible regions of the environment.

However, autonomous driving systems that depend solely on LoS-based sensors face inherent limitations in detecting objects within NLoS regions, which are common in complex urban environments. According to the U.S. Department of Transportation, traffic accidents at intersections resulted in 12,036 fatalities in 2022, highlighting the critical challenge of limited situational awareness in such scenarios \cite{NHTSA2022child}. LoS sensors are incapable of detecting pedestrians or obstacles obstructed by buildings and fences at intersections, posing significant safety risks. To address this limitation, there is a need for perception methods capable of accurately recognizing pedestrians and obstacles in NLoS environments.

\begin{figure}[t!]
	\centering
	\centering
	\includegraphics[width=\linewidth]{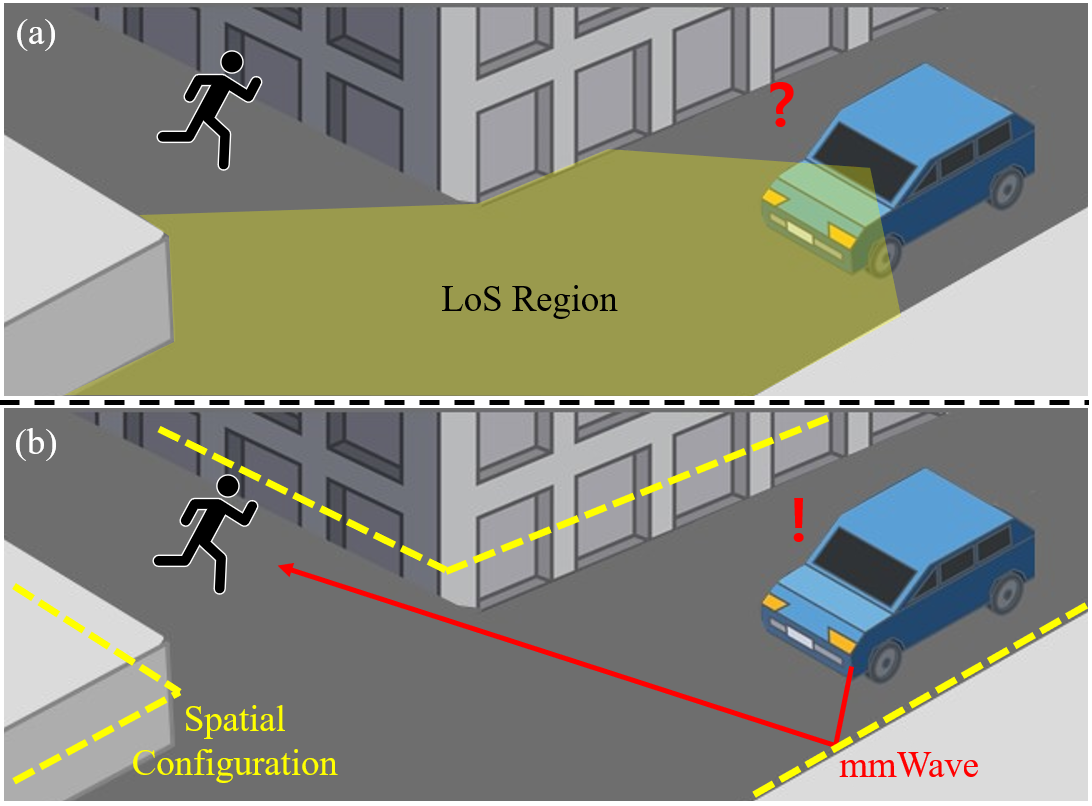}
	\vspace{-0.7cm}
	\caption{\textbf{Illustration of the results using the proposed method in an intersection composed of backroads.} (a) Using only LoS sensors, it is not possible to detect objects in the NLoS region, (b) The proposed method allows for estimating the position of NLoS objects.}
	\label{fig:accidentsituation}
\end{figure}

In response to the limitations of existing solutions, various researches have explored NLoS observation through Vehicle-to-Everything (V2X) communication as a potential solution \cite{rebsamen2012utilizing}. For instance, cooperative perception is proposed as a method utilizing Vehicle-to-Vehicle (V2V) communication, wherein vehicles within a platoon exchange information about their perceived surroundings to improve safety and driving efficiency \cite{kim2013cooperative, liu2013motion, kim2014multivehicle}. However, these solutions are often impractical in environments lacking proper infrastructure and incur substantial deployment and maintenance cost.

As a result, a more generalizable approach to NLoS perception is needed, one that involves independent observation and analysis of occluded environments. This approach requires sensors operating in frequency bands capable of reflecting and diffracting signals, such as mmWave radar. mmWave radar leverages the reflectivity to detect objects in NLoS regions and estimate their positions.

Nevertheless, using 2D radar PCD for spatial inference presents several challenges. 2D radar PCD is often sparse, noisy, and prone to positional distortion due to multi-path reflections. Additionally, interpreting 2D radar PCD requires substantial spatial information to analyze reflection paths. Environments with multiple reflectors, such as T-junctions, pose significant challenges for precise spatial inference.

To overcome these limitations, this paper proposes the approach that integrates 2D radar PCD with front camera images to observe and analyze NLoS regions. While front camera images alone cannot provide precise 3D spatial estimates or directly capture NLoS areas, they offer high-resolution visual information that aids in analyzing road structures and layouts. By leveraging the complementary strengths of both sensors, the proposed framework of this paper infers the structure of occluded spaces and estimates pedestrian locations within NLoS regions is proposed, as shown in Fig. \ref{fig:accidentsituation}.

The proposed approach extracts road layout information from front camera images, serving as a foundation for interpreting 2D radar PCD to infer spatial details. A ray-tracing technique is then used to correct distorted positional data in the 2D radar PCD. Filtering and clustering methods remove noise and enhance spatial inference accuracy. The effectiveness of the framework is evaluated using real-world datasets from a radar-camera system mounted on a vehicle, demonstrating its ability to localize pedestrians in NLoS environments in practical driving scenarios
The main contributions of this research are as follows:  
\begin{itemize}  
	\item A novel framework for NLoS pedestrian localization at T-junctions using mmWave radar and road layout inferred from camera images is proposed.
	\item A method for interpreting 2D radar PCD based on front camera images to achieve accurate spatial inference is proposed.  
	\item The localization performance is validated using real-world outdoor NLoS datasets from a radar-camera system mounted on a vehicle.
\end{itemize}

\section{Related Works}\label{sec:RelatedWorks}

\subsection{Camera-based road layout estimation model} 


In the context of road environment perception, camera-based Bird’s Eye View (BEV) transformation and road layout estimation models have garnered significant attention \cite{li2022bevformerlearningbirdseyeviewrepresentation, peng2023bevsegformer}. Cameras offer a distinct advantage in extracting structures such as lanes, road boundaries, and intersections due to their high-resolution visual information. Recently, deep learning-based techniques for BEV transformation have been widely adopted in the realm of autonomous driving research. Philion \etal proposed the Lift-Splat-Shoot model, which processes multi-camera images to reconstruct the 3D environment and generate a BEV-style road layout \cite{philion2020lift}. This model integrates multi-view information to provide comprehensive spatial representations. Qureshi \etal introduced a deep learning-based approach that estimates the BEV layout of urban driving scenarios from a single image \cite{mani2020monolayout}. By employing adversarial feature learning and multi-channel semantic occupancy grids, this method can infer occluded regions and reconstruct a plausible scene layout in real-time. Liu \etal utilized a transformer-based model to estimate road layouts, including NLoS regions occluded by obstacles, from monocular camera images \cite{liu2024monocular}.


\begin{table}[t!]
	\centering
	\caption{Comparison of NLoS Pedestrian Localization Methods.}
	\label{tab:relatedworks}
	\resizebox{\columnwidth}{!}{
			\begin{tabular}{c|c|cc|cc}
					\hline
					\multirow{2}{*}{Methods}  & Input & \multicolumn{2}{c|}{Object} & Sensor & Reflector  \\
					\cline{3-4}
					 & type & NLoS & Dynamic & fusion &  estimation \\  \hline\hline
					
					\begin{tabular}[c]{@{}c@{}}Chen \etal \\\cite{chen2022non}\end{tabular}  & Signal & \checkmark &  &  &  \\ 
					\hline
					
					\begin{tabular}[c]{@{}c@{}}Pham \etal \\\cite{pham2023multipath}\end{tabular} & Signal & \checkmark & \checkmark &  &  \\
					\hline
									
 					\begin{tabular}[c]{@{}c@{}}Palffy \etal \\\cite{palffy2022detecting}\end{tabular}& \begin{tabular}[c]{@{}c@{}}PCD, \\ Image\end{tabular} & \checkmark & \checkmark  & \checkmark &  \\ 
					\midrule[0.9pt]
					
					\begin{tabular}[c]{@{}c@{}}\textbf{Proposed} \\\textbf{Method}\end{tabular}& \begin{tabular}[c]{@{}c@{}}PCD, \\ Image\end{tabular} & \checkmark & \checkmark & \checkmark & \checkmark \\ 
					
					\hline
			\end{tabular}
		}
\end{table}

\subsection{mmWave radar based NLoS object detection}

Recent research on detecting NLoS objects utilizing the reflective properties of mmWave radar has been actively pursued \cite{zhu2023non, fan2019moving}. Chen \etal proposed a multipath reflection model using MIMO radar and estimated the Time of Arrival for each path through the Matrix Pencil algorithm to infer the position of NLoS objects \cite{chen2022non}. However, this method was only validated in an indoor environment, raising concerns regarding its applicability in outdoor and more complex scenarios. Pham \etal introduced a Bayesian-based multipath selection technique to reduce localization ambiguities in NLoS situations \cite{pham2023multipath}. Despite this improvement, their method does not explicitly reconstruct the spatial environment. Palffy \etal proposed a radar-camera sensor fusion technique for pedestrian localization in NLoS environments, employing a particle filter to estimate the position of pedestrians occluded by vehicles \cite{palffy2022detecting}. However, their approach only detects occluded regions without estimating reflectors, limiting its ability to utilize radar points generated by multiple reflections.

To address these limitations, this paper introduces the method that fuses front-view camera data with mmWave radar, enabling more accurate NLoS pedestrian localization. The key differences between the proposed approach and previous research are summarized in Table \ref{tab:relatedworks}.

\begin{figure*}[t!]
	\centering
	\centering
	\includegraphics[width=\linewidth]{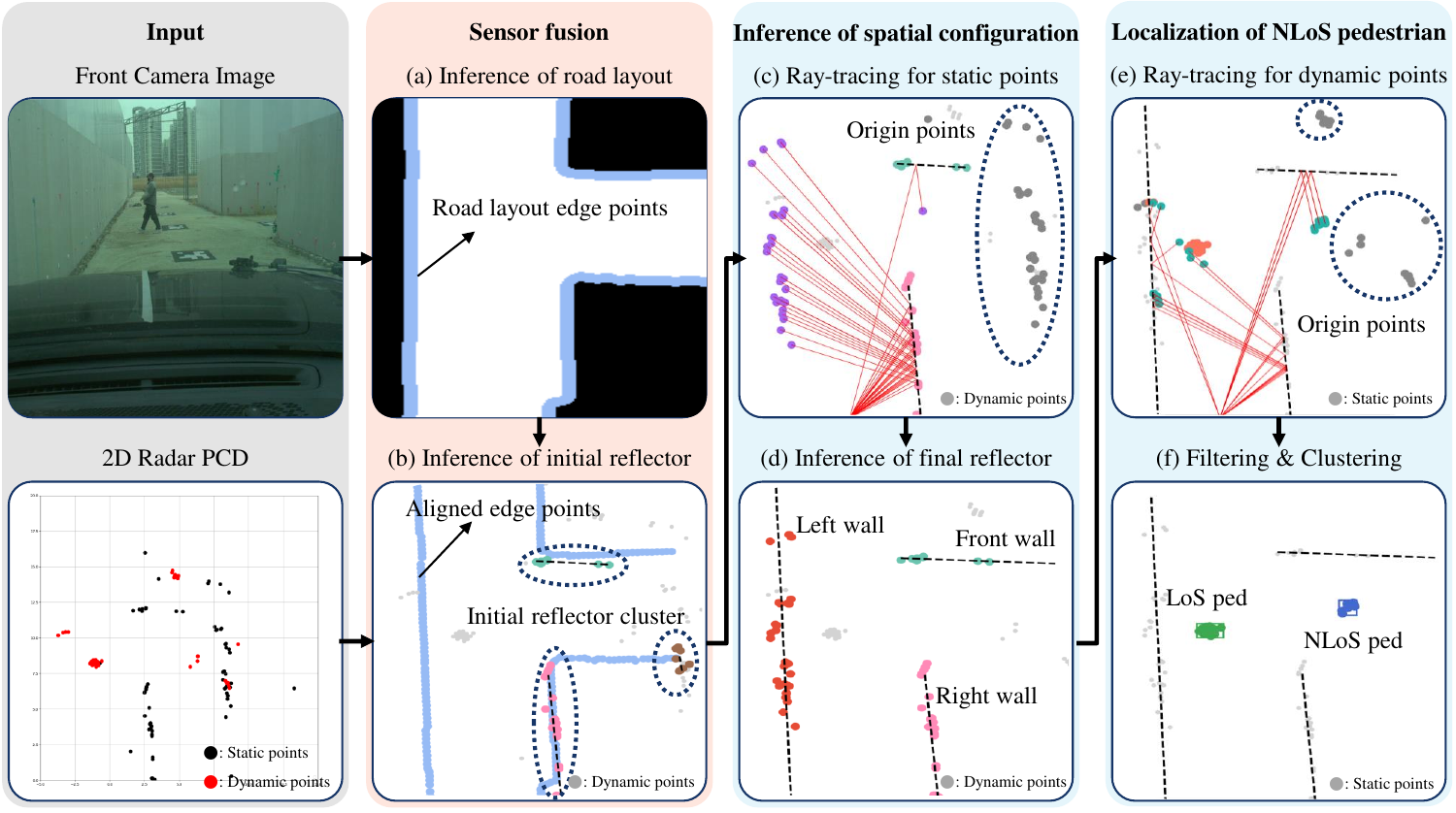}
	\caption{\textbf{The overall framework for NLoS pedestrian localization and the results of each algorithm block.}}
	\label{fig:flowchart}
\end{figure*}



\section{Problem Definition}\label{sec:ProblemDefinition}
Accurately localizing NLoS pedestrians requires analyzing mmWave signal reflection paths, but inferring spatial details from radar PCD alone presents significant challenges. To overcome this limitation, the proposed method incorporates the front camera image to enhance spatial inference, with the radar’s static points in PCD interpreted based on the front camera image.

The proposed method involves a series of steps for spatial inference. First, the road layout is extracted from the front camera image, denoted as $I_{\text{cam}}$, using a model $F_t$ that transforms $I_{\text{cam}}$ into a BEV layout \cite{liu2024monocular}. Based on this road layout, the radar's static points in PCD, denoted as $S = \{ s_1, s_2, \dots, s_n \}$, are interpreted to extract the spatial configuration $L$. This process is expressed as:
\begin{align}
    L = f(S \mid F_t(I_{\text{cam}})),
\end{align}


where $f$ represents the function that interprets the set $S$ based on the road layout extracted from $I_{\text{cam}}$, yielding the spatial configuration $L$.

Next, a dynamic set of radar points in PCD, denoted as $D$, is defined from the radar PCD. Utilizing the spatial configuration $L$, the set $D$ is analyzed to estimate the position of the pedestrian, denoted as $X_{\text{pred}}$. This process is formulated as:
\begin{align}
    X_{\text{pred}} = g(D \mid L),
\end{align}
where $g$ is the function that estimates the pedestrian's position by analyzing the set $D$ based on the spatial configuration $L$. The function $g$ interprets $D$ in the context of $L$, resulting in the predicted position of the pedestrian, $X_{\text{pred}}$.

Finally, the objective of this method is to minimize the absolute error between the predicted pedestrian position $X_{\text{pred}}$ and the ground truth position $X_{\text{GT}}$. The optimization problem is formulated as:
\begin{align}
\label{eq:ae}
    X^*_{\text{pred}} = \arg \min_{X_{\text{pred}}} |X_{\text{pred}} - X_{\text{GT}}|.
\end{align}
The goal is to find the predicted position $X_{\text{pred}}$ that minimizes the absolute error, thereby achieving the most accurate localization of pedestrians in the NLoS region. This process will be further elaborated in Section \ref{sec:Experiments}.

\section{NLoS Pedestrian Localization Pipeline through Fusion of Image \& PCD\label{sec:ProposedMethod}}
To observe NLoS regions, it is essential to utilize waves with reflective and diffractive properties, and by analyzing these properties, the location of NLoS pedestrians can be inferred. This suggests that the results of such analysis may vary depending on the points of reflection and diffraction, ultimately demonstrating a strong dependence on the spatial configuration. However, accurately analyzing spatial information using a single sensor presents significant limitations.

While mmWave radar PCD provides precise distance measurements, it is constrained by sparse observation data and positional distortion due to multipath reflections. Conversely, camera images offer interpretable information for LoS regions but suffer from relatively inaccurate depth estimation and are unable to detect NLoS objects.

To address these challenges, this section proposes a sensor fusion-based spatial analysis pipeline that integrates the quantitative distance data from mmWave radar with the qualitative visual information from a camera. Based on this fusion approach, we introduce a method for estimating the location of pedestrians in NLoS regions, as shown in Fig. \ref{fig:flowchart},

\subsection{Inference of spatial configuration}\label{sec:spatial}


 mmWave radar is capable of distinguishing between static and dynamic objects by analyzing the Doppler effect. In this context, the observation of \( S \) refers to the set of static points obtained from the radar PCD. Given the mounting position of the radar in this experiment, these static points can be considered as observations of reflectors.
The challenge, however, arises from the fact that static objects can be observed through both direct and reflected paths, as formulated by the following expression:
\begin{align}
    S = S_{\text{direct}} \cup S_{\text{reflect}},
\end{align}
where \( S_{\text{direct}} \) represents the static points observed through the direct path, and \( S_{\text{reflect}} \) represents the static points observed through the reflected path.

\begin{algorithm}[t!]
	\caption{Inference of spatial configuration}
	\label{alg:spatial_configuration}
	\KwIn{$I_{cam}$: Front camera image, $S = \{ s_1, s_2, \dots, s_n \}$: Set of static points from 2D Radar PCD}
	\KwOut{$L$: Final set of linear regressions representing reflectors}
	\BlankLine
	
	\tcp{\textit{Inference of road layout}}
	$I_{layout} \gets F_t(I_{cam})$\;
	\BlankLine

        \tcp{\textit{Extract edge points in $I_{layout}$}}
	$I_{edge} \gets \text{ExtractEdges}(I_{layout})$\;
        $x_{\text{wall}} \gets X_{\text{scale}} \cdot (u - o_x) + x_{\text{offset}} \quad \text{where } (u, v) \in I_{\text{edge}}$\;
        $y_{\text{wall}} \gets Y_{\text{scale}} \cdot (o_y - v) + y_{\text{offset}}$\;
        $W \gets \{ w_j = (x_{\text{wall}}, y_{\text{wall}}) \mid j = 1, 2, \dots, m \}$\;
	\BlankLine
    
        \tcp{\textit{Alignment of $W$ \& radar PCD}} 
        $W_{\text{near}} \gets \{ w_j \mid \|s_i - w_j\| < \epsilon \}$\;
        $\theta, T \gets \arg\min_{\theta, t_x, t_y} \sum_{w_j \in W_{\text{near}}} \min_{s_i \in S} \| s_i - w_j \|$\;
        $W_{\text{aligned}} \gets R(\theta) W_{\text{near}} + T$\;
        \BlankLine
        
        \tcp{\textit{Initial interpretation $S$ using $W_{aligned}$}}
        $C \gets \{ C_1, C_2, ..., C_k \}$; \tcp{\textit{DBSCAN on $W_{aligned}$}}
	\BlankLine
	\ForEach{$C_{ref} \in C$}{
            $S_j \gets \{s_i | \exists w \in W_j, \ \|s_i - w\| < \delta \}$\;
            $l_j  \gets LinearRegression(S_j)$; 
	}
	\BlankLine
	\tcp{\textit{Ray tracing for reflected static points}}
	\ForEach{$r \in S_{\text{reflect}}$}{
        $\alpha, \beta \gets \text{FindIntersection}(r, L)$\; 
        $r'_x \gets 2\alpha(r_y - \beta) + r_x \cdot \frac{\alpha^2 + 1}{\alpha^2 + 1 - r_x}$\;
        $r'_y \gets 2 \left( \alpha \left( \frac{r'_x + r_x}{2} \right) + \beta \right) - r_y$\;
        $S_{\text{relocated}} \gets S_{\text{relocated}} \cup \{r'\}$\;
	}
	\BlankLine
	\tcp{\textit{Inference of final spatial configuration}}
        $S_{final} \gets S_{\text{direct}} \cup S_{\text{relocated}}$\;
        \ForEach{$C_{ref} \in C$}{
        $S'_j \gets \{ s'_i \mid s'_i \in S_{\text{final}}, \exists w \in W_j, \ \|s'_i - w\| < \delta \}$\;
        $l_j  \gets LinearRegression(S'_j)$; 
	}
	\Return $L \gets \{ l_j | j \in \{1, 2, ..., k\} \}$\;
\end{algorithm}

In this context, the coordinates of points observed through the reflected path do not correspond to the actual position of the object but instead contain distorted positional information. Thus, a method for interpreting these points is required. To address this, the present method proposes an algorithm that interprets \( S \) through \( I_{\text{layout}} \) to infer the spatial configuration as summarized in Algorithm \ref{alg:spatial_configuration}.

\subsubsection{Inference of road layout}\label{sec:spatial-1}

Recent research on road layout inference models utilizing camera images have been actively pursued. In this research, a model that can infer the road layout, including previously unseen road areas, from monocular camera images is required. One such model, which satisfies this requirement, takes \( I_{\text{cam}} \) as input and infers the road layout image \( I_{\text{layout}} \), as formulated follows:
\begin{align}
    I_{\text{layout}} &= F_{t}(I_{\text{cam}}).
\end{align}

\subsubsection{Extract edge points in $I_{layout}$}\label{sec:spatial-2}

The road layout image \( I_{\text{layout}} \) is divided into two regions: the drivable space and the undrivable space, which are represented in a binary occupancy map format. Specifically, the pixel values of the drivable region are set to 255, while those of the undrivable region are set to 0. To extract the boundaries of the road from \( I_{\text{layout}} \), we detect edges by examining the value changes between adjacent pixels, where the Euclidean distance between pixels is 1. This process is applied to pixels \( (u, v) \) in \( I_{\text{layout}} \) where the pixel value is 255. By doing so, we extract a set of points, \( I_{\text{edge}} \), that represent the boundaries of the road.

The coordinates of the points in the extracted \( I_{\text{edge}} \) are initially expressed in pixel units, necessitating a conversion process into radar coordinate units (meters). The transformation of \( m \) transformed image points, \( W = \{ w_j = (x_{\text{wall}}, y_{\text{wall}}) | j \in \{1, 2, \dots, m\} \} \), is given by the following equations:
\begin{align}  
    x_{\text{wall}} &= (u - o_x) \cdot X_{\text{scale}} + x_{\text{offset}},  \\  
    y_{\text{wall}} &= (o_y - v) \cdot Y_{\text{scale}} + y_{\text{offset}}. 
\end{align}
where \( X_{\text{scale}} \) and \( Y_{\text{scale}} \) are the scale factor values, which are hyperparameters. \( x_{\text{offset}} \) and \( y_{\text{offset}} \) are the correction values that account for the difference in mounting positions between the radar and the camera. \( o_x \) represents the horizontal central coordinate of the front camera in \( I_{\text{layout}} \), and \( o_y \) represents the vertical bottom coordinate.

\subsubsection{Alignment of $W$ \& radar PCD}\label{sec:spatial-3}
The set \( S \) contains observations of multiple reflectors, and without distinguishing each reflector as an individual instance, it is impossible to infer accurate spatial information. However, due to the sparse nature of the observation data, it is difficult to determine whether the points originate from a single structure or multiple reflectors. Thus, relying solely on \( S \) may lead to incorrect spatial inferences. To classify the reflectors as distinct instances in \( S \), it is necessary to interpret \( S \) based on the qualitative form of \( W \). However, since \( W \) provides inaccurate quantitative location information, we must reinterpret \( S \) through \( W \) first.

To achieve this, we first identify \( W_{\text{near}} \), the set of points in \( W \) that are within a certain distance \( \epsilon \) from each point \( s \in S \), as formulated below:
\begin{align}  
    W_{\text{near}} &= \{ w_j \mid \|s_i - w_j\| < \epsilon \}.
\end{align}

Next, we convert \( W_{\text{near}} \) into an aligned set of road boundary points in the radar coordinate system, denoted as \( W_{\text{aligned}} \), as follows:
\begin{align}  
    W_{\text{aligned}} &= R(\theta) W_{\text{near}} + T,
\end{align}
where \( R(\theta) \) is the rotation matrix and \( T = (t_x, t_y) \) is the translation vector. To optimize \( \theta \) and \( T \), we employ a method that minimizes the sum of Euclidean distances between points in \( W_{\text{near}} \) and \( S \). The optimization process is formulated as:
\begin{align}  
    \theta, T \gets \arg\min_{\theta, t_x, t_y} \sum_{w_j \in W_{\text{near}}} \min_{s_i \in S} \| s_i - w_j \|.
\end{align}

The optimization process adjusts the values of \( \theta \), \( t_x \), and \( t_y \) using the Nelder-Mead method, which allows us to obtain the optimal transformation parameters \cite{10.1093/comjnl/7.4.308}. Once the optimization is complete, the obtained rotation matrix \( R(\theta) \) and translation vector \( T \) are applied to the entire set of \( W_{\text{near}} \) to transform it into the radar-aligned set \( W_{\text{aligned}} \).

\subsubsection{Initial interpretation of $S$ using $W_{\text{aligned}}$}\label{sec:spatial-4}

To accurately infer the environment based on the observations from \( S \), it is essential to identify the points corresponding to the reflectors. To achieve this, we first apply the DBSCAN algorithm \cite{ester1996density} to \( W_{\text{aligned}} \) to partition the data into distinct clusters, each representing a potential reflector, denoted as \( \{ W_j \mid j \in \{1, 2, \dots, k\} \} \).

Subsequently, to identify the points in \( S \) that correspond to specific reflectors, we extract the subset \( S_j \) from \( S \) where the distance to any point \( w \in W_j \) is within a defined threshold \( \delta \), as formulated below:
\begin{align}
    S_j &= \{ s_i \mid \exists w \in W_j, \ \| s_i - w \| < \delta \}.
\end{align}

Next, a linear regression model is applied to the points in \( S_j \) to derive the line segment that best represents the geometry of each reflector. The regression process utilizes the Least Squares Method (LSM) to estimate the slope and y-intercept of the line representing each cluster \( S_j \), denoted by \( \alpha \) and \( \beta \), respectively. The resulting linear equation \( l_j \) for each reflector is expressed as:
\begin{align}
    l_j : y = \alpha x + \beta.
\end{align}

Through this process, we obtain a set of linear models, \( L = \{ l_1, l_2, \dots, l_k \} \), each representing an inferred reflector in the environment.

\subsubsection{Ray tracing for reflect static points}\label{sec:spatial-5}


The estimated set \( L \) contains spatial information inferred as \( S_{\text{reflect}} \), and the process of transforming \( S_{\text{reflect}} \) to infer the environment is required. In this method, based on the previously inferred set \( L \), we differentiate \( S_{\text{reflect}} \) from \( S_{\text{direct}} \) and apply ray-tracing techniques to adjust the position of \( S_{\text{reflect}} \).

The ray-tracing model traces the path from the origin \( O \) of the Ego-vehicle to a point \( s \in S_{\text{reflect}} \), identifying the intersection points where this path intersects the line segments \( L \), which represent the boundaries of reflectors. Among these intersection points, the one closest to the origin is selected as \( q \). The point \( r = (r_x, r_y) \in S_{\text{reflect}} \) is then symmetrically reflected across the line \( l_j : y = \alpha x + \beta \), which passes through \( q \). The recalibrated points, \( S_{\text{relocated}} = \{r' = (r_x', r_y')\} \), are computed as follows:
\begin{align}
    r'_x &= \frac{2 \alpha (r_y - \beta) + r_x}{\alpha^2 + 1} - r_x,
\end{align}
\begin{align}
    r'_y &= 2 \left( \alpha \frac{(r'_x + r_x)}{2} + \beta \right) - r_y.
\end{align}

\subsubsection{Inference final spatial configuration}\label{sec:spatial-6}
The set of points \( S_{\text{final}} \), reconstructed to align with the actual road environment, is defined as follows:
\begin{align}
    S_{\text{final}} = S_{\text{direct}} \cup S_{\text{relocated}}.
\end{align}
Similar to the procedure in Section \ref{sec:spatial-4}, linear regression is once again performed on \( S_{\text{final}} \) to ultimately derive the final spatial information \( L \).

\subsection{Localization of NLoS pedestrian}\label{sec:localization}

A pedestrian is observed as a dynamic object represented by a set of points \( D \) in the 2D radar PCD. To estimate the actual position of a NLoS pedestrian, ray tracing, as described in Section \ref{sec:spatial}, is employed to infer the real-world position of the pedestrian. 
However, due to the potential presence of clutter and noise in the 2D radar PCD, which may arise from high reflectivity, localization is conducted through a series of filtering and clustering processes. The algorithm is outlined as follows.

\subsubsection{Ray tracing of dynamic points}\label{sec:localization-1}

The set \( D \) represents the observed points of moving objects, which include the set of points acquired through direct paths, denoted as \( D_{\text{direct}} \), and the set of points observed after being reflected by reflectors (walls), denoted as \( D_{\text{reflect}} \). The set of points in \( D_{\text{reflect}} \) that are corrected via ray tracing is referred to as the set of relocated points, \( D_{\text{relocated}} \). Thus, the complete set of dynamic points, including the relocated points, can be expressed as follows:
\begin{equation}
    D' = D_{\text{direct}} \cup D_{\text{relocated}}.
    \label{eq:direct}
\end{equation}

\subsubsection{Filtering \& clustering}\label{sec:localization-2}

As the observation distance increases in the 2D radar PCD, the detection error distance becomes larger, and noise arises due to hardware limitations and environmental factors, which subsequently reduce the accuracy of NLoS pedestrian detection. To address these issues, filtering and clustering processes are applied.

The set of relocated points, \( D_{\text{relocated}} \), can be represented as the union of points in the LoS region \( D'_{\text{los}} \) and the set of points in the NLoS region \( D'_{\text{nlos}} \), as formulated below:
\begin{equation}
D_\text{relocated} = D'_{\text{los}} \cup D'_{\text{nlos}}.
\end{equation}

Pedestrians in the LoS region are observed through both \( D_{\text{direct}} \) and \( D'_{\text{los}} \). However, since \( D'_{\text{los}} \) is located further away than \( D_{\text{direct}} \), its reliability is lower. Thus, the position of LoS objects can be more accurately estimated using only \( D_{\text{direct}} \). Therefore, for LoS objects, only the more reliable \( D_{\text{direct}} \) is used, and based on Equation \ref{eq:direct}, the points ultimately used for pedestrian localization, \( D_{\text{final}} \), are defined as follows:
\begin{equation}
D_{\text{final}} = D_{\text{direct}} \cup D'_{\text{nlos}}.
\end{equation}

Subsequently, DBSCAN is applied to \( D_{\text{final}} \) to estimate the pedestrian's position \( X_{\text{pred}} \) while simultaneously removing residual noise, as formulated below:
\begin{equation}
X_{\text{pred}} = \text{DBSCAN}(D_{\text{final}}).
\end{equation}

\section{Dataset}\label{sec:Datasets}
\subsection{Test bed \& data acquisition vehicle}


In real-world road environments, data collection is constrained by safety, and NLoS areas cannot be directly observed. To address this challenge, a testbed with dimensions of \( 53.5 \, \text{m} \times 33.5 \, \text{m} \), encompassing various road conditions, was constructed. To estimate pedestrian positions in the NLoS regions, a camera equipped with a fisheye lens was installed at a height of 7$m$ at the center of the intersection to provide a BEV of the entire experimental area. 
For data collection, various sensors were mounted on an SUV, with sensor positions chosen based on those found in actual vehicles. For NLoS object detection, two 77 GHz mmWave radar sensors were placed on the front-right side of the vehicle, and a front camera was mounted at the rearview mirror position to differentiate between NLoS and LoS conditions of objects. Additionally, a LiDAR was integrated to calibrate the sensors and evaluate the inferred spatial configuration. Data from all sensors were collected at 100 ms intervals. Further details can be found in Jeon \etal \cite{jeon2024non}.

\subsection{Data for road layout model}

A total of 2,931 images were utilized as training data, and 2,200 images were used for validation to train and evaluate the road layout model. The training dataset consisted of annotated images captured while the ego-vehicle was operating within the testbed. Initially, the obtained BEV images were annotated by marking the ego-vehicle's position with a bounding box. For road layout annotation, segmentation was performed, where drivable pixels were represented by a value of 255, and undrivable pixels were represented by a value of 0, resulting in the creation of an occupancy map. Subsequently, based on the annotated position of the ego-vehicle, a Region of Interest was defined. A 1400 × 1400 pixel area was then cropped from the occupancy map, centered around the ego-vehicle's bounding box, to create the corresponding ground truth image.


\section{Experimental Results}\label{sec:Experiments}

To evaluate the performance of the proposed method, validation was carried out in four distinct scenarios, assuming that the ego-vehicle is stationary at a T-junction. The scenarios are categorized into two main cases based on the position of the ego-vehicle: \textbf{B1} refers to the case where the ego-vehicle is located on the Left Main Road, with no wall opposite the T-junction, while \textbf{B2} corresponds to the case where the ego-vehicle is situated on the Branch Road, with a wall opposite the T-junction. Additionally, four specific scenarios were defined based on the movement of pedestrians approaching the intersection. \textbf{S1} represents the scenario where two NLoS pedestrians approach from the right. \textbf{S2} involves one NLoS pedestrian approaching from the right, while one LoS pedestrian moves away from the ego-vehicle. \textbf{S3} describes the case where two NLoS pedestrians approach from the left, and one NLoS pedestrian approaches from the right. Lastly, \textbf{S4} refers to the scenario where two NLoS pedestrians approach from the left, while one LoS pedestrian moves away from the ego-vehicle. Each of these scenarios consists of 60 to 90 frames, allowing for a comprehensive evaluation of the proposed method in various environments and confirming its ability to accurately localize NLoS pedestrians in T-junction scenarios.

\begin{figure}[t!]
	\centering
	\centering
	\includegraphics[width=\linewidth]{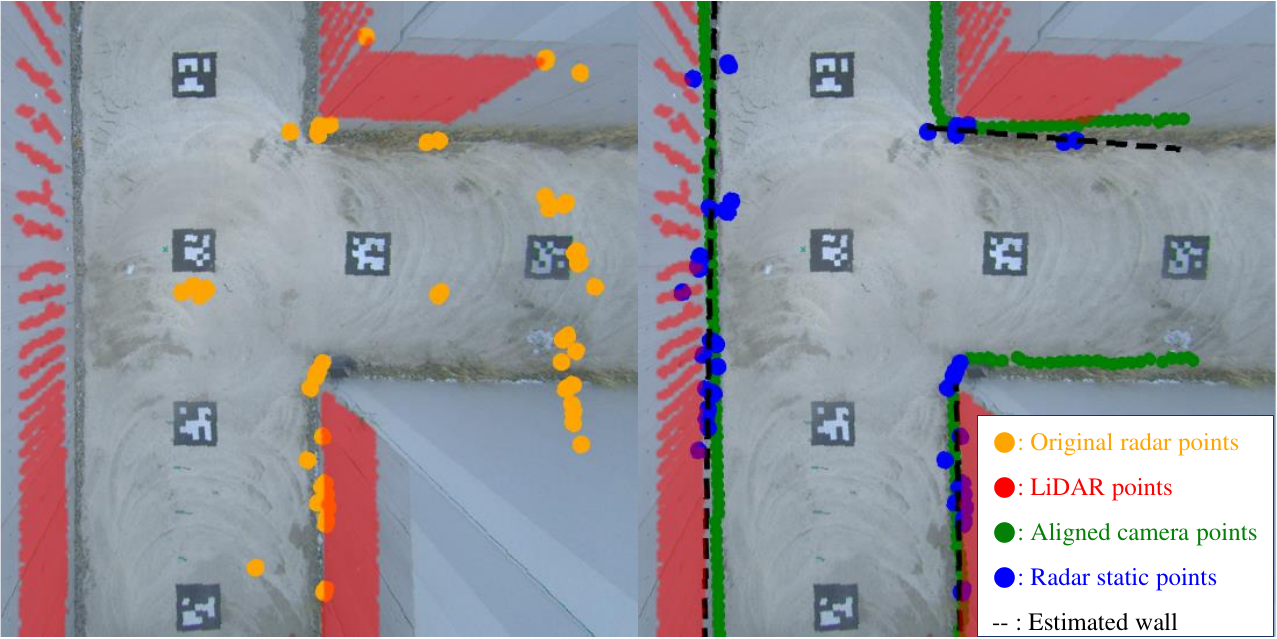}
	\caption{\textbf{Comparison of walls estimated Using LiDAR PCD, aligned camera points, and radar PCD.} (Left) The original radar static points are sparse and contain noise. (Right) The walls estimated using the proposed method are closely aligned with the LiDAR points, showcasing the effectiveness of the approach.}
	\label{fig:SpatialRes}
\end{figure}


\begin{figure*}[t!]
	\centering
	\centering
	\includegraphics[width=\textwidth]{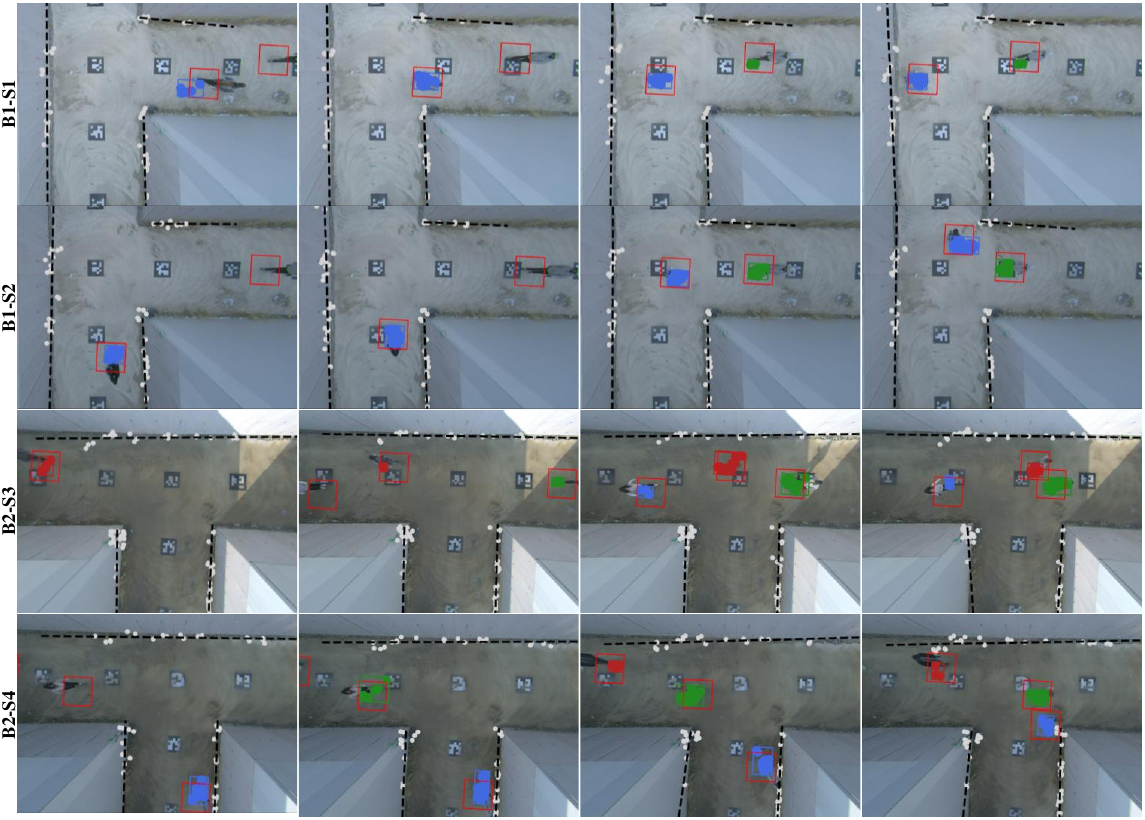}
		\vspace{-0.7cm}
	\caption{\textbf{Qualitative results in each scenarios.}  The figure is organized into four scenarios, displayed from top to bottom. Each scenario is evaluated over time, with frames progressing left to right. Additionally, the distance between each AR mark is 4m and speed of pedestrians is approximately 1.5 $m/s$.}
	\label{fig:QualRes}
\end{figure*}

\subsection{Analysis of 2D Radar PCD using camera data}



By analyzing the spatial information of 2D radar PCD alongside data from a front camera, the accuracy of spatial inference can be significantly enhanced. As shown in Fig. \ref{fig:SpatialRes}, radar points corresponding to the front wall are sparsely distributed. When inferring the spatial structure using only radar points, it becomes difficult to determine whether each point originates from a single reflector or multiple reflectors. To address this, the road layout from the front camera can be used to classify the radar points. This classification enables more effective interpretation of the radar points, improving the overall interpretability of the radar PCD.

\subsection{Evaluation of the spatial configuration inference model}

\begin{table}[t!]
\centering
\caption{Evaluation of the Spatial Configuration Inference Model.}
\vspace{-0.3cm}
\label{tab:AngleQuanResLocal}
\resizebox{\columnwidth}{!}{
    \begin{tabular}{c|c|c|cc|c}
            \hline
            \multicolumn{2}{c|}{Scenarios}  & \multirow{2}{*}{Sensors} & \multicolumn{2}{c|}{Difference [\textdegree]} & Max. Diff.   \\
            \cline{1-2} \cline{4-5}
            Sites & Cases  &  & F-R & F-L & with LiDAR \\  \hline\hline
            
            \multirow{6}{*}{\textbf{B1}}&\multirow{3}{*}{\textbf{S1}} & Radar only & 83.77 & 79.59 & 7.74\\ 
            &&Proposed    & 84.94 &83.64 & 3.69\\
            \cline{3-6}
            && LiDAR (GT)  & 87.38 & 87.33 & -\\ 
            \cline{2-6}

            &\multirow{3}{*}{\textbf{S2}}   &Radar only    & 77.41 & 79.10 & 9.99\\
            &&Proposed    & 85.72 & 85.00 &2.34\\
            \cline{3-6}
            && LiDAR (GT) & 87.40 & 87.34 & -\\ 
            \hline

            \multirow{6}{*}{\textbf{B2}}&\multirow{3}{*}{\textbf{S3}}   & Radar only & 94.2 & 76.49 &10.89\\ 
            &&Proposed    & 84.94 &83.64 &3.74\\
            \cline{3-6}
            && LiDAR (GT)  & 87.44 & 87.38 & -\\ 
            \cline{2-6}

            &\multirow{3}{*}{\textbf{S4}}   &Radar only    & 89.17 & 79.96 & 7.44 \\ 
            &&Proposed    & 85.72 & 85.00 & 2.4\\
            \cline{3-6}
            && LiDAR (GT) & 87.44 & 87.40 & -\\ 
            \hline            
        \end{tabular}
    }
\end{table}

In this paper, the pedestrian localization algorithm was evaluated to verify the performance of the proposed spatial configuration inference model. Various sensor configurations, including LiDAR (used as the ground truth), Radar-only, and camera-radar fusion (proposed), were tested. Furthermore, to minimize calibration errors of each sensor, the angular differences between the front wall and the right wall, as well as between the front wall and the left wall, were computed for evaluation. The Front Wall-Right Wall Angular Difference (F-R) and Front Wall-Left Wall Angular Difference (F-L) are defined as follows:
\begin{align}
    \text{F-R} &= W^{\text{Front}}_a - W^{\text{Right}}_a, \\
    \text{F-L} &= W^{\text{Front}}_a - W^{\text{Left}}_a,
\end{align}
where \( W_a \) represents the angle formed with the x-axis.


As presented in Table \ref{tab:AngleQuanResLocal}, the proposed method exhibited an average error of 2–4 degrees relative to the LiDAR-based measurements, which served as the ground truth, demonstrating high spatial inference accuracy. The evaluation results underscore the effectiveness of the proposed method in reducing the angular difference between F-R and F-L in comparison to the Radar-only method. For instance, in the B1-S1 case, the F-R difference for the Radar-only method was 83.77°, and the proposed method improved this to 84.94°, representing a reduction of 1.40\%. In the B1-S2 case, the F-R difference for the Radar-only method was 77.41°, and the proposed method improved it to 85.72°, reflecting a 10.70\% enhancement. Similarly, in the B2-S3 and B2-S4 scenarios, the proposed method achieved improvements in F-R, with reductions of 10.00\% and 3.89\%, respectively.

\begin{table}[t!]
\centering
\caption{Evaluation of pedestrian localization.}
\vspace{-0.3cm}
\label{tab:QuanResLocal}
\scriptsize
\resizebox{\columnwidth}{!}{
    \begin{tabular}{c|c|c|cc|c}
            \hline
            \multicolumn{2}{c|}{Scenarios}  & \multirow{2}{*}{Methods} & \multicolumn{3}{c}{Localization Error [AE]}   \\
            \cline{1-2} \cline{4-6}
            Sites & Cases &  & NLoS & LoS & AVG  \\  \hline\hline
            
            \multirow{4}{*}{\textbf{B1}}&\multirow{2}{*}{\textbf{S1}} & Radar only & 2.13 & 1.85 & 1.81 \\ 
            && Proposed &  0.86 & 0.26 & \textbf{0.40}  \\
            \cline{2-6}

            &\multirow{2}{*}{\textbf{S2}}   & Radar only & 1.33 & 1.62 & 1.61 \\ 
            &&Proposed & 0.33 & 0.37 & \textbf{0.36} \\
            \hline

            \multirow{4}{*}{\textbf{B2}}&\multirow{2}{*}{\textbf{S3}}   & Radar only & 0.98 & 1.73 & 1.55 \\ 
            &&Proposed & 0.29 & 0.43 & \textbf{0.37} \\
            \cline{2-6}

            &\multirow{2}{*}{\textbf{S4}}   & Radar only & 0.64 & 1.69 & 1.61 \\ 
            &&Proposed & 0.43 & 0.44 & \textbf{0.44} \\
            \hline
            
        \end{tabular}
    }
\end{table}
In all four scenarios, the proposed method showed a 6.5\% improvement in F-R and a 4.6\% improvement in F-L angular differences compared to the Radar-only method. The average angular differences for F-R and F-L were reduced from 5.52° and 6.86° to 2.08° and 3.04°, respectively, highlighting the proposed method's ability to enhance spatial configuration inference. The Radar-only method, due to the radar's sparse left wall points and the radar's position on the front-right side of the vehicle, showed greater error in F-L localization. However, the proposed method improved F-L localization accuracy, outperforming Radar-only’s F-R estimation. This demonstrates the effectiveness of radar-camera fusion, particularly in NLoS environments, where camera-derived road layout information compensates for the radar’s limited view, improving robustness for autonomous driving.

\subsection{Evaluation of pedestrian localization model}
The pedestrian localization performance of the proposed method was quantitatively evaluated based on the Absolute Error (AE), which is defined in Equation \ref{eq:ae}.

In this method, considering the presence of multiple pedestrians, the predicted pedestrian position \( X_{\text{pred}} \) was matched with the closest actual pedestrian position \( X_{\text{GT}} \) to calculate the AE. If only one pedestrian was predicted in a given frame, the error between the predicted position and the closest ground truth position was calculated. In cases where multiple pedestrians were predicted, the average AE for each frame was used as the performance metric. The AE was computed separately for three categories: NLoS pedestrian estimation, LoS pedestrian estimation, and the average of both. The experimental results are summarized in Table \ref{tab:QuanResLocal}. In the B1-S1 scenario, the AE was recorded as 0.40 m, and in the B1-S2 scenario, it was 0.36 m. In the B2-S3 scenario, the AE was 0.37 m, and in the B2-S4 scenario, it was 0.44 m. The inference process takes 0.07 seconds per frame, resulting in a frame rate of 13.7 FPS on a GTX 1080 GPU. The proposed method demonstrated consistent localization performance across different scenarios, with the AE consistently within the 0.44 m range for all scenarios.
In the \textbf{B1} branch experiment, NLoS pedestrian localization is achieved through the radar→front wall→pedestrian→front wall→radar reflection path. However, in the Radar Only method, due to the sparsity of the radar data, the inference of the front wall is inaccurate, which results in a degradation of NLoS pedestrian localization performance. On the other hand, the proposed method enhances the performance by effectively addressing this issue, demonstrating improved localization accuracy, as shown in Fig. \ref{fig:QualRes}.

\section{Conclusions}\label{sec:Conclusion}

This paper introduced a novel sensor fusion framework that combined 2D radar PCD with front camera images to achieve accurate pedestrian localization in NLoS environments. The proposed method leveraged front camera images to enhance the interpretation of 2D radar PCD, facilitating spatial inference and the localization of NLoS pedestrians at T-junctions. The proposed method showed a 6.5\% improvement in F-R angular difference and a 4.6\% improvement in F-L angular difference, compared to the radar only method. Furthermore, when the localization performance was evaluated using AE, the proposed method demonstrated an accuracy of at least 1.18 $m$ across all scenarios, outperforming the radar only method, with an average accuracy of 1.25 $m$. Additionally, all estimated results of the proposed method were within 0.44 $m$.These results demonstrate the effectiveness of the proposed method in improving pedestrian localization in NLoS environments.


\addtolength{\textheight}{-12cm}   

\bibliographystyle{IEEEtran}
\bibliography{reference,IEEEabrv}

\end{document}